\documentclass[lettersize,journal]{IEEEtran}
\usepackage{amsmath,amsfonts}
\usepackage{algorithmic}
\usepackage{array}
\usepackage[caption=false,font=normalsize,labelfont=sf,textfont=sf]{subfig}
\usepackage{textcomp}
\usepackage{stfloats}
\usepackage{url}
\usepackage{verbatim}
\usepackage{graphicx}
\def\BibTeX{{\rm B\kern-.05em{\sc i\kern-.025em b}\kern-.08em
    T\kern-.1667em\lower.7ex\hbox{E}\kern-.125emX}}
\usepackage{balance}
\usepackage{cite}
\usepackage{newfloat}
\usepackage{listings}

\usepackage{epsfig}
\usepackage{amsmath}
\usepackage{makecell}

\usepackage{subcaption}
\usepackage{multirow}
\usepackage{booktabs}
\newcommand{\R}{\mathbb{R}}

\usepackage{colortbl}
\usepackage{xcolor}
\usepackage{amssymb}
\usepackage{pifont}
\definecolor{green}{RGB}{0, 133, 21} 
\definecolor{red}{RGB}{0.988,0.914,0.914}
\definecolor{gray}{RGB}{128, 128, 128}
\definecolor{improvecolor}{RGB}{112, 173, 71}
\newcommand{\tablestyle}[2]{\setlength{\tabcolsep}{#1}\renewcommand{\arraystretch}{#2}\centering\footnotesize}
\def\eg{\emph{e.g}} 
\def\ie{\emph{i.e}}

\begin{document}
\title{Taming Generative Synthetic Data for X-ray \\ Prohibited Item Detection}

\author{Jialong Sun, Hongguang Zhu, Weizhe Liu, Yunda Sun, Renshuai Tao and Yunchao Wei

\thanks{
Jialong Sun, Weizhe Liu, Renshuai Tao and Yunchao Wei are with Institute of Information Science, Beijing Jiaotong University, Beijing, 100044, China (email: \{sunjialong, liuweizhe, rstao, yunchao.wei\}@bjtu.edu.cn).

Hongguang Zhu is with Faculty of Data Science, City University of Macau, China 
(email: zhuhongguang1103@gmail.com).

Yunda Sun is with Nuctech Company Limited, Beijing, 100083, China (email: sunyunda@nuctech.com).

}}

\markboth{Journal of \LaTeX\ Class Files,~Vol.~18, No.~9, September~2020}%
{How to Use the IEEEtran \LaTeX \ Templates}

\maketitle

\begin{abstract}
Training prohibited item detection models requires a large amount of X-ray security images, but collecting and annotating these images is time-consuming and laborious. To address data insufficiency, X-ray security image synthesis methods composite images to scale up datasets. However, previous methods primarily follow a two-stage pipeline, where they implement labor-intensive foreground extraction in the first stage and then composite images in the second stage. Such a pipeline introduces inevitable extra labor cost and is not efficient. 
In this paper, we propose a one-stage \textbf{X}-ray security image \textbf{syn}thesis pipeline (\textbf{Xsyn}) based on text-to-image generation, which incorporates two effective strategies to improve the usability of synthetic images. 
The \textbf{C}ross-\textbf{A}ttention \textbf{R}efinement (\textbf{CAR}) strategy leverages the cross-attention map from the diffusion model to refine the bounding box annotation. The \textbf{B}ackground \textbf{O}cclusion \textbf{M}odeling (\textbf{BOM}) strategy explicitly models background occlusion in the latent space to enhance imaging complexity.
To the best of our knowledge, compared with previous methods, Xsyn is the first to achieve high-quality X-ray security image synthesis without extra labor cost. 
Experiments demonstrate that our method outperforms all previous methods with 1.2\% mAP improvement, and the synthetic images generated by our method are beneficial to improve prohibited item detection performance across various X-ray security datasets and detectors. Code is available at \url{https://github.com/pILLOW-1/Xsyn/}.
\end{abstract}


\section{Introduction}

\IEEEPARstart{A}{utomatic} prohibited item detection~\cite{tifs1,tifs2,tifs3,tifs4,tifs5,tifs6,01_isaac23evaluation,03_web21xray,04_bhowmik21energy,05_isaac20multiview, gaus25anomaly, bhowmik22subcomponent} aims to detect all contraband from a single X-ray security image. Training such models usually requires a large amount of annotated data, but both collecting and annotating X-ray security images are time-consuming and laborious, resulting in a high labor cost for obtaining well-annotated images.
For example, collecting a single image from X-ray security inspection equipment can take up to one minute, and multiple rounds of iterative labeling by professional annotators further increase time costs.

To reduce the cost of collecting hand-annotated X-ray security images, utilizing synthetic data has emerged as an effective way. 
Previous X-ray image synthesis methods mainly utilize two methods to synthesize images: Threat Image Projection-based (TIP-based) synthesis~\cite{bhowmik2019good,duan2023rwsc} and Generative Adversarial Network-based (GAN-based) synthesis~\cite{goodfellow2014GAN,zhu2020data,zhao2018gan,yang2019data,li2021gan}. 1) TIP-based synthesis involves fusing the prohibited item with the background image through morphological operations~\cite{bhowmik2019good} or an image fusion neural network~\cite{duan2023rwsc}. However, it either requires laborious mask annotation for foreground extraction or time-consuming Foreground Threat Image (FTI) collection for fusion network training. 2) GAN-based synthesis enriches the foreground diversity by adopting GAN~\cite{goodfellow2014GAN} to generate prohibited items with varying poses and shapes. 
However, training GAN on foreground images also brings inevitable extra labor cost on data collection and annotation (\eg, FTI collection~\cite{zhu2020data}, trimap~\cite{yang2019data}, and semantic label~\cite{li2021gan}). 

\begin{figure}
    \centering
    \includegraphics[width=1\linewidth]{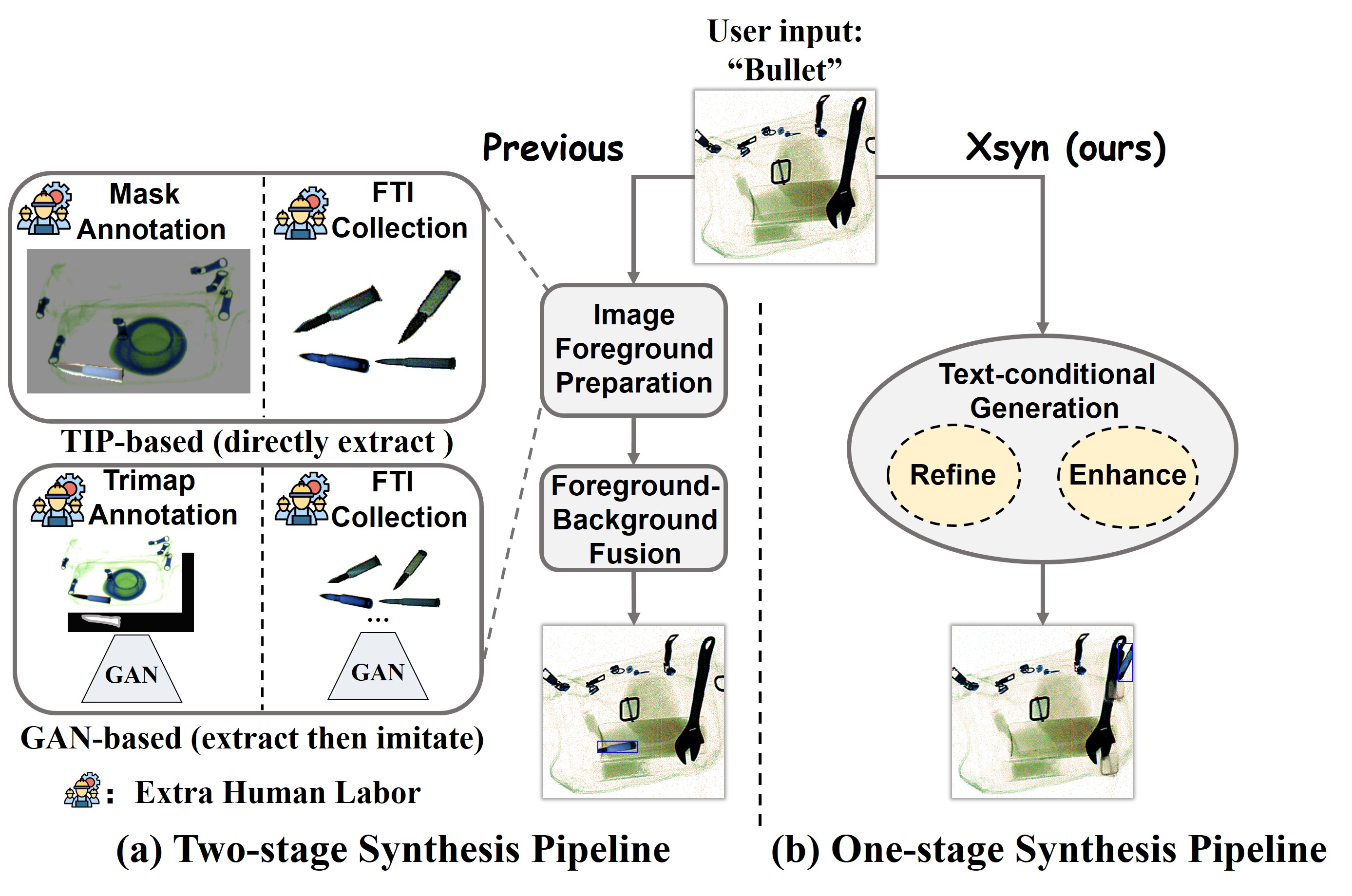}
    \caption{Analysis of existing X-ray security image synthesis methods. Previous two-stage synthesis methods introduce inevitable labor cost in the first stage (\eg, foreground preparation process), which hinders the efficiency of the whole synthesis pipeline. In contrast, Xsyn is a simple and effective one-stage synthesis pipeline, which can automatically refine the synthetic annotation and enhance the synthetic complexity, thereby generating high-quality synthetic data and eliminating extra labor costs.}
    \label{fig:previous_methods_analysis}
\end{figure}

As shown in Figure~\ref{fig:previous_methods_analysis}, we analyze existing X-ray security image synthesis methods, observing that there is one common limitation in previous methods: \textit{they all suffer from inevitable extra labor (\eg, FTI collection and annotation)}. We argue that this limitation stems from the fact that previous methods primarily follow a two-stage synthesis pipeline, where the first stage involves extracting foregrounds for the second synthesis stage, thus introducing inevitable extra labor shown in Figure~\ref{fig:previous_methods_analysis} (a). For instance, TIP-based methods directly extract image foregrounds, and GAN-based methods imitate these foregrounds on the basis of extraction. Therefore, the question arises: \textit{Can we achieve high-quality X-ray security image synthesis without extra labor?}

In this paper, we propose a simple and effective one-stage \textbf{X}-ray security image \textbf{syn}thesis (\textbf{Xsyn}) pipeline to eliminate extra labor cost.  
The basic idea is illustrated in Figure~\ref{fig:previous_methods_analysis} (b). 
Our method is based on the text-grounded inpainting pipeline, which requires no extra labor cost and can generate high-quality X-ray security images by bridging the generative power of the diffusion model and the perception capability of SAM~\cite{kirillov2023SAM}.
Specifically, we fine-tune the layout-to-image diffusion model through text-grounded inpainting training and then inpaint X-ray security images by providing grounding conditions (\eg, bounding boxes with class names). 
To refine synthetic annotations of the generated X-ray security images, we propose \textbf{C}ross-\textbf{A}ttention \textbf{R}efinement (\textbf{CAR}), which refines the bounding box through the cross-attention map from the diffusion model.
By designing a median point sampling strategy based on the most class-discriminative part of the cross-attention map, we augment the bounding box prompt and input it to SAM, thus obtaining precise position prediction.
Considering the common background occlusion in real-world baggages, we further introduce \textbf{B}ackground \textbf{O}cclusion \textbf{M}odeling (\textbf{BOM}) to enhance synthetic complexity, which explicitly models background occlusion in the latent space of the diffusion model. We propose to automatically search the background occluder and then fuse the background occluder with the foreground parts of the latent at the end of the denoising process. 
With the above strategies, our synthesis method can generate high-quality X-ray security images without labor-intensive cost. These synthetic images can be used to train prohibited item detection models, supplementing real images.
To 
summarize, our contributions are threefold:
\begin{itemize}
    \item We propose Xsyn, a simple and effective one-stage synthesis pipeline in the X-ray security domain. To the best of our knowledge, Xsyn is the first to achieve high-quality X-ray security image synthesis without incurring additional labor-intensive foreground preparation.

    \item We present two effective strategies to enhance the usability of synthetic data. The CAR strategy automatically refines the synthetic image annotations, and the BOM strategy explicitly models the background occlusion in X-ray security images to enhance their imaging complexity.

    \item Experiments on public X-ray security datasets demonstrate that the generated images from our synthesis pipeline are beneficial to improve prohibited item detection performance.

\end{itemize}

\section{Related Work}
\label{sec:related_work}

\textbf{X-ray Security Image Synthesis.} Prohibited item detection models require a large amount of data. Considering the training need, X-ray security image synthesis~\cite{bhowmik2019good, duan2023rwsc, zhu2020data, yang2019data, zhao2018gan, li2021gan} has emerged as an effective way to deal with data insufficiency. It can mainly be categorized into two ways: TIP-based synthesis~\cite{bhowmik2019good,duan2023rwsc} and GAN-based synthesis~\cite{zhu2020data,zhao2018gan,yang2019data,li2021gan}. 1) TIP-based synthesis augments X-ray imagery datasets by superimposing prohibited items onto available X-ray security baggage images. For example, TIP~\cite{bhowmik2019good} blends isolated threat objects onto benign X-ray images through multistage morphological operations and composition. RWSC-Fusion~\cite{duan2023rwsc} trains an end-to-end region-wise style-controlled fusion network that superimposes prohibited items onto normal X-ray security images to synthesize realistic composite images. 2) GAN-based synthesis aims to directly generate prohibited items.  Yang~\cite{yang2019data} proposes to extract prohibited items with KNN-matting~\cite{chen2013knn} and improve CT-GAN~\cite{wei2018ct-gan} for prohibited item generation. Li~\cite{li2021gan} presents a GAN-based method for synthesizing X-ray security images with multiple prohibited items by establishing a semantic label library. Zhu~\cite{zhu2020data} propose an improved Self-Attention GAN (SAGAN)~\cite{zhang2019sagan} to generate diverse X-ray images of prohibited items and integrate them with background images. However, the aforementioned methods all suffer from inevitable extra labor costs, including time-consuming FTI collection~\cite{ bhowmik2019good,duan2023rwsc,zhu2020data}, mask~\cite{bhowmik2019good}, trimap~\cite{zhao2018gan,yang2019data}, and semantic label~\cite{li2021gan} annotation cost. In contrast to previous methods, our method removes extra labor costs and can generate high-quality X-ray security images through an automatic synthesis pipeline.

\textbf{Generative Data Synthesis for Detection. }A series of methods~\cite{chen2023geodiffusion,ge2022dalle_det,wacv2024_data_aug,zhao2023x-paste,wang2024detdiffusion} have utilized generative models for detection data generation in the natural image domain, and can mainly be divided into two manners~\cite{chen2023geodiffusion}: copy-paste synthesis~\cite{ge2022dalle_det,zhao2023x-paste} and layout-to-image (L2I) generation~\cite{li2023gligen,wang2024detdiffusion,chen2023geodiffusion}. 1) Copy-paste synthesis aims to generate separate foreground objects and fuse them with background images. Ge~\cite{ge2022dalle_det} decouples detection data generation into foreground object mask generation and background image generation through DALL-E~\cite{ramesh2021dall-e}. Zhao~\cite{zhao2023x-paste} leverages CLIP~\cite{radford2021clip} and Stable Diffusion~\cite{ldm} to obtain images with accurate categories for copy-paste synthesis. However, copy-paste synthesis requires separate foreground image generation, which can bring inevitable extra labor costs in the X-ray security domain.
2) The L2I methods, on the other hand, directly generate the whole image with objects from the layout instruction (\eg, bounding boxes with object categories), avoiding the need to generate foregrounds separately. To achieve better controllable generation, GLIGEN~\cite{li2023gligen} integrates a novel gated self-attention mechanism into text-to-image diffusion models for better layout control. GeoDiffusion~\cite{chen2023geodiffusion} further translates geometric conditions into text prompts to generate high-quality detection data. To eliminate the extra labor cost, our method is built upon layout-to-image generation, but extends it into text-grounded inpainting to deal with the background distribution discrepancy in the X-ray security domain, and distinctively proposes two effective strategies to improve the usability of generated images.

\section{Preliminary}
\label{preliminary}
Latent Diffusion Model~\cite{ldm} is a kind of diffusion model that performs the diffusion process in the latent space for text-to-image generation. Specifically, 
given a noisy latent $\textbf{z}_t \in \R^{H^{'} \times W^{'} \times C}$ at each timestep $t\in\{0,...,T-1\}$, 
a denoising UNet~\cite{unet} $\epsilon_\theta(\cdot)$ is trained to recover its clean version $\textbf{z}_0$ by predicting the added noise, and the training objective can be formulated as follows:
\begin{equation}
    \mathcal{L}_{LDM} = \mathbb{E}_{\textbf{z}_0,\epsilon\sim\mathcal{N}(0,1),t}\|\epsilon - \epsilon_\theta(\textbf{z}_t, t, \textbf{c})\|^2
    \label{equ:ldm}
\end{equation}
where $\epsilon$ is the added random Gaussian noise and \textbf{c} is the generalized condition. For text-to-image generation, \textbf{c} is the text prompt which will be encoded by a pre-trained CLIP~\cite{radford2021clip} text encoder. For layout-to-image generation, \textbf{c} further incorporates the grounding condition (\eg, bounding boxes with categories).

Eq.~\ref{equ:ldm} can be further reformulated to support inpainting tasks. Specifically, given an inpainting mask $\textbf{m}$ and the input image, the input image latent $\textbf{z}_0^{input}$ can be extracted by a pre-trained Vector
Quantized Variational AutoEncoder (VQ-VAE)~\cite{van2017vqvae}, and its masked version $\textbf{z}_0^{mask}$ is the multiplication of $\textbf{z}_0^{input}$ and $\textbf{m}^{resize}$, where $\textbf{m}^{resize}$ is obtained by resizing $\textbf{m}$ to the latent size. Based on~\cite{li2023gligen}, the input for UNet is expanded as
%
$\textbf{z}_t^{inpaint}=Concat(\textbf{z}_t, \textbf{z}_0^{mask}, \textbf{m}^{resize})$, which is fed into Eq.~\ref{equ:ldm} to replace $\textbf{z}_t$ for inpainting training. Then, at each sampling step $t$, the noisy latent $\textbf{z}_t$ is updated as follows before denoising:
\begin{equation}
    \textbf{z}_{t} = \textbf{z}_{t+1} * (1 - \textbf{m}^{resize}) + \textbf{z}_{t}^{input} * \textbf{m}^{resize}
\end{equation}
where $\textbf{z}_t^{input}$ is the noisy version of $\textbf{z}_0^{input}$.

\begin{figure}
    \centering
    \includegraphics[width=1\linewidth]{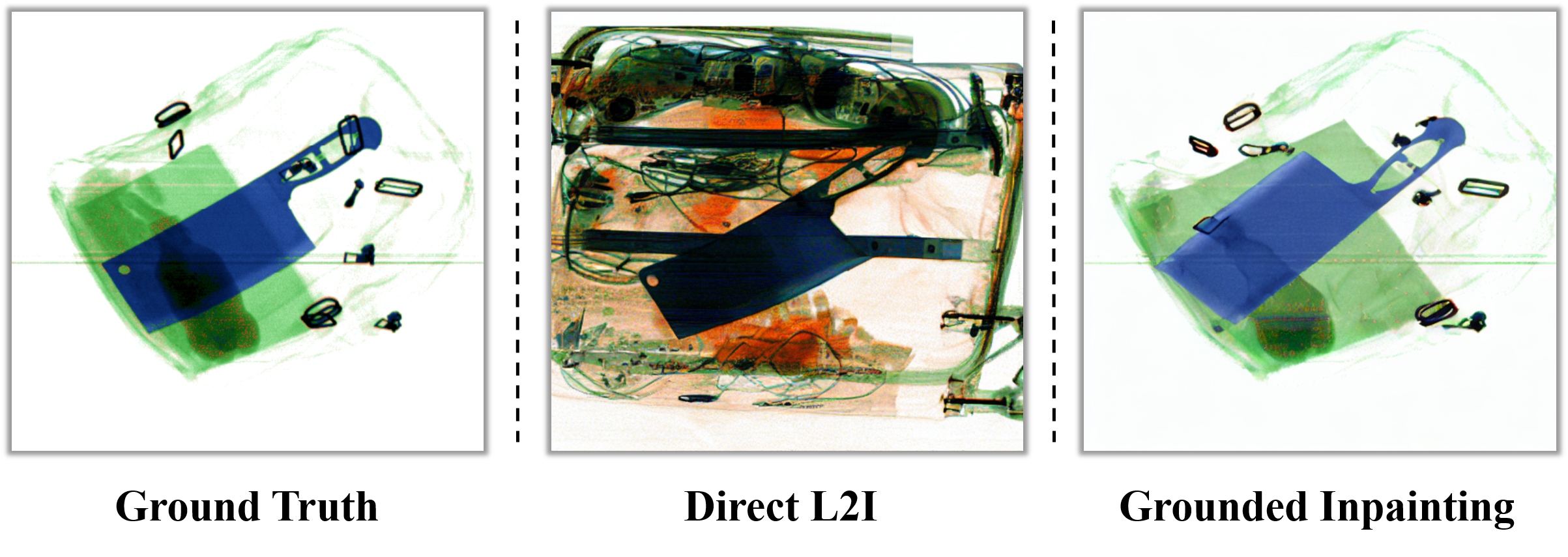}
    \caption{Qualitative comparisons between L2I generation and grounded inpainting. The background of the L2I-generated image (middle) differs a lot from the real-world baggage (left), which may hinder the detection performance. Therefore, we choose grounded inpainting (right) to retain the background.}
    \label{fig:inpaint_l2i_compare}
\end{figure}

\section{Methodology}
\label{method}

\subsection{Generation Pipeline}
\label{sec:pipeline}

Previous L2I methods~\cite{chen2023geodiffusion,wang2024detdiffusion} in the natural image domain directly use the generated images as synthetic data. However, we find that such an approach is not feasible in the X-ray security image domain since the background of the generated image is uncontrollable and its distribution deviates significantly from that of the real background, as shown in Figure~\ref{fig:inpaint_l2i_compare}. To avoid the above problem, we base the generation pipeline on text-grounded inpainting. 

In general, given an X-ray security image $I \in \R^{H \times W \times 3}$, a text prompt $Y$, and a grounding condition $G$, the text-grounded inpainting process can be formulated as a function $ I^* = F(I, Y, G)$. The grounding condition $G = \{(e_i, l_i)\}_{i=1}^M$, where $e_i$ represents the textual description of the object (\eg, class name), and $l_i=[x_{i,1}, y_{i,1}, x_{i,2}, y_{i,2}]$ denotes the $i$-th grounding box (\ie, top-left and bottom-right coordinates). The output is an image with the grounding region being repainted, as specified by the text prompt $Y$ and the grounding condition $G$. 

To generate a new X-ray security image, we design two kinds of grounding conditions $G_{mod}$ and $G_{add}$ based on the image annotation $L$, where $ L = \{(c_i, b_i)\}_{i=1}^N $, $c_i$ represents the class name, and $b_i$ represents the $i$-th annotation box, sharing the same format as the grounding box. Specifically, we first let $G_{mod} = L$ so that we can reuse the annotation and modify the geometry of the original prohibited items (\eg, shape and pose). 
To add a new prohibited item to an image, we first use SAM to segment all elements within the image. Subsequently, we discard the two largest masks by area to prevent out-of-boundary generation. Because they typically correspond to the background and the whole baggage region.
Then we select an idle region $l_b$ from the rest masks randomly, and $l_b$ satisfies the following criterion,
\begin{equation}\label{lb_criterion}
    l_b \in \{ 
        l \in S \mid dis(l, b_i) < d, i = 1,2,\ldots, N
        \},
\end{equation}
where $S=\{ s_k \}_{k=1}^K$, $s_k$ is the bounding box of the $k$th object segmented by SAM in image $I$, $dis(\cdot,\cdot)$ measures the IoU between two bounding boxes and $d$ is the pre-defined threshold. In practice, boxes that are too small will be filtered out. Then we select a category $c_b$ for $l_b$ from a class group which corresponds to specific region areas 
and let $e_b = c_b$ to obtain $G_{add} = \{ (e_b, l_b)\}$. By concatenating the class names as the text prompt, we get $Y_{mod} = Concat(\{c_i\}_{i=1}^N)$ and $Y_{add} = \{e_b\}$. Finally, we can generate a new image in two different ways as follows,
\begin{equation} \label{I1}
\begin{aligned}
    I^*_{mod} &= F(I, Y_{mod}, G_{mod}),  \\
    I^*_{add} &= F(I, Y_{add}, G_{add})
\end{aligned}
\end{equation}

Therefore, we can construct two variants of synthetic data using Eq.~\ref{I1}, named Xsyn-M and Xsyn-A, respectively. This generation pipeline has two advantages. First, it does not require any extra labor cost (\eg, FTI collection) compared with previous synthesis methods. Second, it focuses on generating foreground items by altering only a portion of the background, without affecting the overall distribution.

\begin{figure}[t]
    \centering
    \includegraphics[width=1\linewidth]{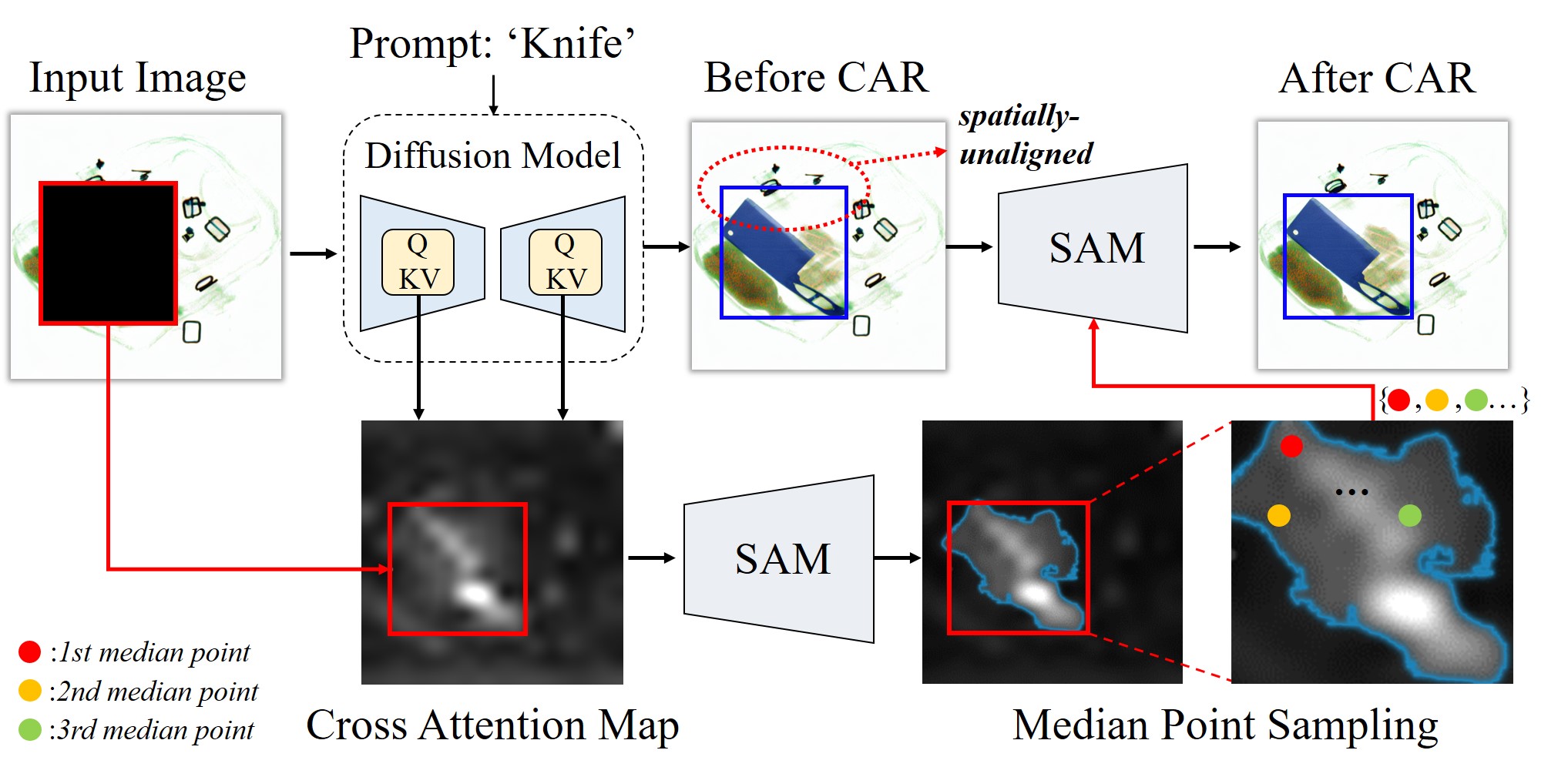}
    \caption{Cross-Attention Refinement. To obtain the spatial-aligned annotation, we leverage SAM to locate the generated prohibited item based on the rich class-discriminative spatial localization information in the cross-attention map. Please see how the bounding box (blue box) of the generated item is refined.}
    \label{fig:scar}
\end{figure}

\subsection{Cross-Attention Refinement}
\label{SCAR}

Because it is hard for the generated item to be tightly within the grounding box, directly using the grounding box as the annotation box to train detection models will lead to performance degradation of downstream tasks. 
Instead of forcing the generative model to generate spatially aligned items, we retain the generated item and propose CAR to refine its location to obtain the aligned annotation.

Given an input X-ray security image, we first inpaint it using the proposed generation pipeline.
Directly using SAM to refine the location by taking the grounding box as input is suboptimal (refer to Table~\ref{tab:ablation_param}), because the background can affect the performance of SAM.
To address the above issue, we step out of the image domain and propose CAR based on the cross-attention map in the diffusion model.
Figure~\ref{fig:scar} shows the process of CAR. For simplicity, we only discuss the refinement process for one generated item. For the generated item corresponding to $g_i= (e_i, l_i) \in G$, we obtain the average cross-attention map $M_i \in \R^{H \times W}$ from the diffusion model for the text token corresponding to $e_i$. 
The CAR process takes as input the generated image $I^*$, the cross-attention map $M_i$, and the grounding location $l_i$. The output is the refined annotation location for the generated item. Specifically, we first obtain the most class-discriminative region $r_i$ by using SAM to segment $M_i$ within $l_i$. To help SAM better locate the generated item based on $l_i$, we then propose a median point sampling strategy to sample points $P_i$ from $l_i$ and combine these points with $l_i$ as prompt $\mathcal{P}_i$ for SAM to locate the generated item, where $\mathcal{P}_i = P_i \cup \{ l_i \}$. 

\begin{figure}[t]
    \centering
    \includegraphics[width=1\linewidth]{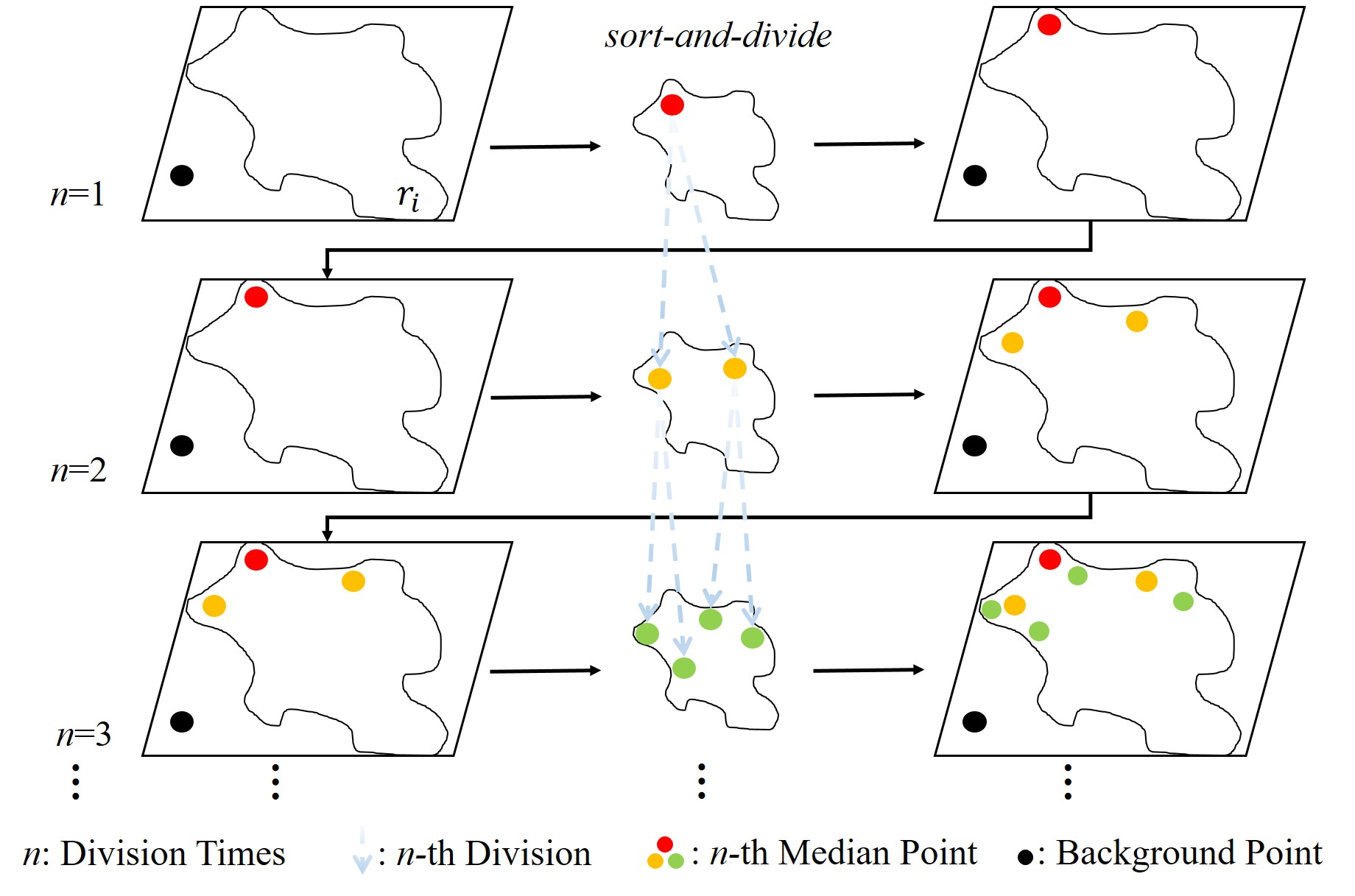}
    \caption{Median Point Sampling. Considering the background in the bounding box may interfere with the refinement, we propose to enhance the localization by sampling median points as foreground points in a recursive manner.  Different colors refer to different division levels.}
    \label{fig:point_sampling}
\end{figure}

\textbf{Median Point Sampling (MPS). }Figure~\ref{fig:point_sampling} depicts the basic idea of median point sampling. We aim to sample foreground points inside $r_i$ and background points outside $r_i$. Specifically, we choose the point with the minimum activation value outside $r_i$ as the background point $p_i^{b}$. To sample foreground points, we first sort all points within $r_i$ by their activation values and choose the median point $p_i^{f_1}$ as the first foreground point. Then we divide $r_i$ into two sub-regions $r_i^{1}$ and $r_i^{2}$, where the activation values in $r_i^{1}$ are all below that of $p_i^{f_1}$, and the activation values in $r_i^{2}$ are all above that of $p_i^{f_1}$. By extension, we perform the same \textit{sort-and-divide} operation on the subsequent sub-regions recursively and gather all the median points as foreground points. Therefore, we obtain the final point set 
$P_i = \{ p_i^{f_1}, p_i^{f_2}, \ldots, p_i^{f_{2^{n-1}}}, \ldots, p_i^{f_{2^{n}-1}}, p_i^{b}\}$ which has $2^{n}-1$ foreground points and one background point in total, where $n$ indicates the division times. For example, the red, orange, and green points in Figure~\ref{fig:point_sampling} are in the 1st, 2nd, and 3rd divisions, respectively. We argue that median points describe the central tendency of data points belonging to the prohibited item, which are less affected by extreme activation values in the cross-attention map. 

Finally, the refinement process uses SAM to segment $I^*$ by taking $\mathcal{P}_i$ as visual prompts and assigns the bounding box of the segmented region to be the annotation box, thus obtaining more precise location prediction for the generated item.
The CAR strategy takes advantage of the segmentation capability of SAM and the cross-attention map of the diffusion model to obtain the refined bounding box annotation. Despite its simplicity, our CAR strategy can achieve automatic annotation refinement that benefits prohibited item detection performance.

\begin{figure}[t]
    \centering
    \includegraphics[width=1\linewidth]{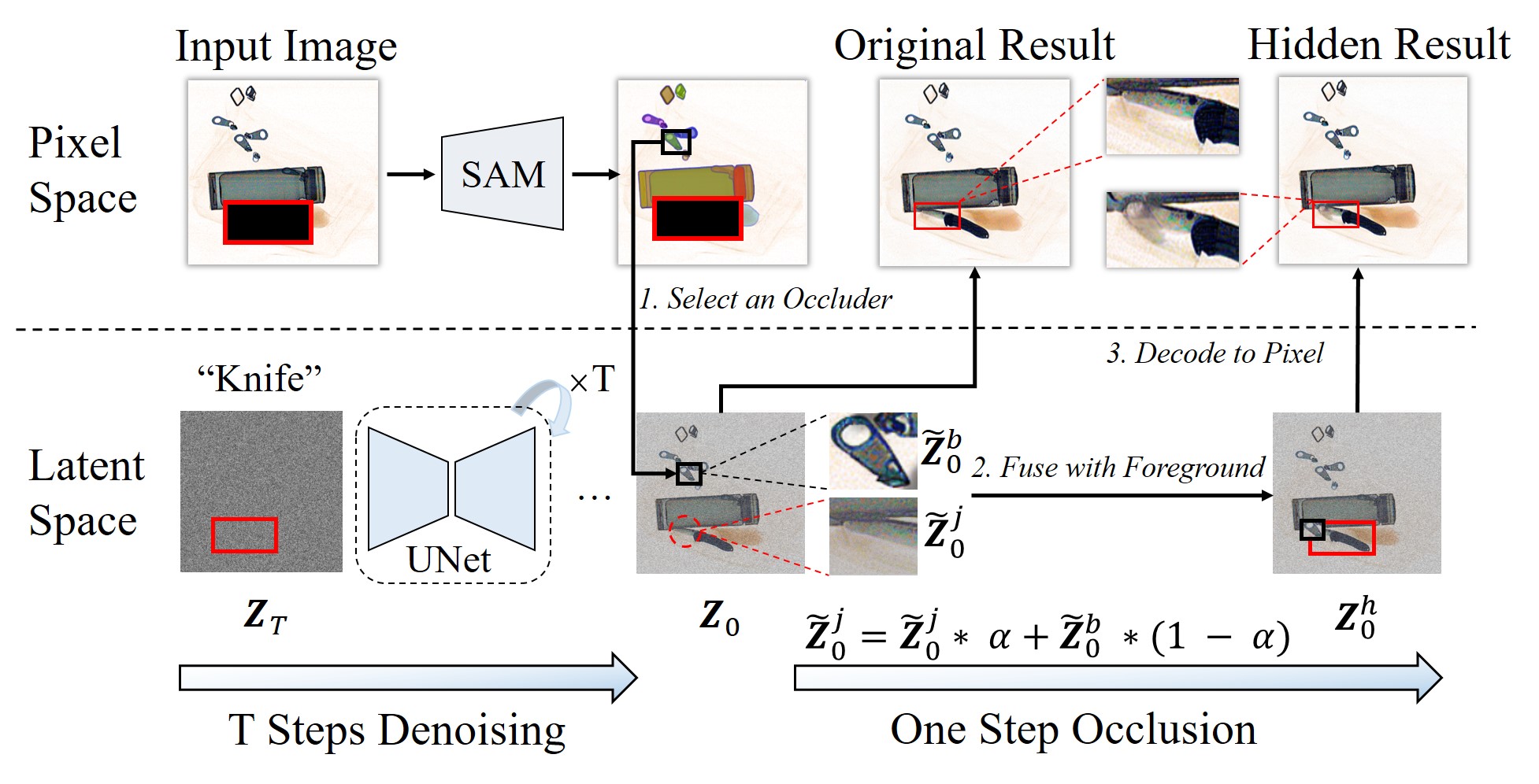}
    \caption{Background Occlusion Modeling. BOM performs occlusion through regional recombination in the latent space. For simplicity, we omit other variables and components of the diffusion model since the whole generation process has been elaborated. }
    \label{fig:BLOM}
\end{figure}

\subsection{Background Occlusion Modeling}
\label{LOM}

The generated prohibited items are too clear, which is inconsistent with complex real-world occlusion scenarios and may induce overfitting problems for detection.
To address the above problem and further enrich the imaging complexity of synthetic images, we simulate the common background occlusion in real baggage by applying background occlusion modeling shown in Figure~\ref{fig:BLOM}, which fuses the specified background region with foreground regions in the latent space to occlude prohibited items.

Specifically, given an input X-ray security image $I$, we first select an occluder from the background in pixel space by using SAM to segment every object in $I$, and use Eq.~\ref{lb_criterion} to determine the location of the occluder\footnote{
Eq.~\ref{lb_criterion} used here is reformed as 
$
    l_o \in \{ 
        l \in S \mid dis(l, b_i) < d, dis(l, l_b) < d, i = 1,2,\ldots, N
        \}
$ if we use $G_{add}$.
}. Next, we adopt the proposed generation pipeline to inpaint $I$ but slightly modify the latent sampling process. As shown in Figure~\ref{fig:BLOM}, a noisy latent $\textbf{z}_T \in \R^{H^{'} \times W^{'} \times C}$ sampled from the standard normal distribution $\mathcal{N}(0,1)$ is passed to the denoising UNet, to obtain the denoised latent $\textbf{z}_0$ after $T$ steps of denoising. If we directly decode $\textbf{z}_0$ to the pixel space, then we will get the original result with no occlusion. To occlude the prohibited item, we perform a weighted recombination of the occluder region and foreground regions in latent space for one more step as follows:
\begin{equation}\label{occlusion_fuse_1}
    \tilde{\textbf{z}}_0^{j} = \tilde{\textbf{z}}_0^{b} * \alpha + \tilde{\textbf{z}}_0^{j} * (1 - \alpha) 
\end{equation}
\begin{equation}\label{occlusion_fuse_2}
    \tilde{\textbf{z}}_0^{j} = Crop(\textbf{z}_0, l_j^{'})
\end{equation}
\begin{equation}\label{occlusion_fuse_3}
    \tilde{\textbf{z}}_0^{b} = Crop(\textbf{z}_0, l_o^{'})
\end{equation}
where $\tilde{\textbf{z}}_0^{j}$ and $\tilde{\textbf{z}}_0^{b}$ is the $j$-th occluded foreground region and the occluder region of $\textbf{z}_0$ respectively. $\alpha$ adjusts the degree of occlusion. $Crop(\cdot)$ represents the process of cropping $\textbf{z}_0$ to the region corresponding to the occluded region $l_j^{'}$ or the occluder region $l_o^{'}$, where $l_o^{'}=[x_{o,1}^{'}, y_{o,1}^{'}, x_{o,2}^{'}, y_{o,2}^{'}]$, and $l_j^{'}$ can be obtained as follows:
\begin{equation}
    l_j^{'}\in\{
    Re(l_j, l_o^{'}) \mid l_j \in G \cup L, j=1, 2, \ldots, M+N
    \}
\end{equation}
where $Re(\cdot)$ first projects $l_j$ to latent space and then perturbs it as follows:
\begin{equation}\label{occlusion_fuse_4}
\begin{aligned}
x_{j,1}^{'} &= Rand(Max(x_{j,1}^{'}-w_o^{'}, 0), x_{j,2}^{'}),
\\
y_{j,1}^{'} &= Rand(Max(y_{j,1}^{'}-h_o^{'}, 0), y_{j,2}^{'}),
\\
 x_{j,2}^{'} &= Min(x_{j,1}^{'} + w_o^{'}, W^{'}),
 \\
 y_{j,2}^{'} &= Min(y_{j,1}^{'} + h_o^{'}, H^{'})
 \end{aligned}
\end{equation}
where $Rand(\cdot)$ randomly samples an integer between the lower bound and the upper bound. $w_o^{'}$ and $h_o^{'}$ is the width and height of $l_o^{'}$ respectively. We let $l_j^{'}=[x_{j,1}^{'}, y_{j,1}^{'}, x_{j,2}^{'}, y_{j,2}^{'}]$ be the $j$-th occluded region. The hidden version of $\textbf{z}_0$ is termed as $\textbf{z}_0^{h}$. Finally, we decode $\textbf{z}_0^{h}$ to pixel space and obtain the hidden result shown in Figure~\ref{fig:BLOM}.   

Through the regional recombination enabled by BOM, the foreground region can be occluded by a random item from the background, which enhances the imaging complexity of synthetic images. It is worth noting that the original result in Figure~\ref{fig:BLOM} is used by CAR to obtain the refined annotation, and we adopt the hidden result as the final synthetic image.

\section{Experiments}
\label{Exp}

\begin{table*}[t]
    \centering
    \caption{Category groups of X-ray security datasets. We split the categories into three groups according to their mean areas for each dataset. The classes in the same group share similar mean areas, and each group belongs to an area interval. Taking PIDray as an example, the area intervals of group1, group2 and group3 are [0, 10000], [10000, 25000] and [25000, \textit{max}] respectively, where \textit{max} is the maximum object area in PIDray. The area boundaries in the table indicate two endpoints dividing the three intervals. }
\label{tab:suppl_category_group}
\tablestyle{4pt}{1.2}\scriptsize
\setlength{\tabcolsep}{12pt}
\begin{tabular}{l|c|c|c|c}
\toprule
Dataset  & Group1 & Group2 & Group3 & Area Boundaries\\ \midrule
PIDray~\cite{zhang2023pidray}  &Lighter, Bullet  & \makecell[c]{Knife, Gun, Powerbank, Wrench, \\ HandCuffs, Baton, Pliers, Scissors, Sprayer} & Hammer & (10000, 25000) \\ \midrule
OPIXray~\cite{wei2020opixray} & Multi-tool\_Knife, Folding\_Knife &  Straight\_Knife, Utility\_Knife & Scissor & (10000, 15000) \\ \midrule
HiXray~\cite{tao2021Hixray} & \makecell{Portable\_Charger\_1, Portable\_Charger\_2, Water, \\ Mobile\_Phone, Cosmetic, Nonmetallic\_Lighter}  & Tablet  & Laptop & (40000, 100000) \\

 \bottomrule
\end{tabular}

\end{table*}

\subsection{Experimental Setups}
\label{exp_setup}
\textbf{Datasets.} We conduct experiments on three widely used X-ray security datasets: PIDray~\cite{zhang2023pidray}, OPIXray~\cite{wei2020opixray}, and HiXray~\cite{tao2021Hixray}. Specifically, PIDray dataset consists of 29,454 training images and 18,220 validation images with bounding box and mask annotations from 12 classes, while OPIXray dataset has 7,109 training images and 1,776 validation images with bounding box annotations from 5 classes. HiXray dataset is composed of 36,295 training images and 9,069 validation images with bounding box annotations from 8 classes.

\textbf{Implementation Details. }
\textit{Generation.} We base the generation pipeline on GLIGEN~\cite{li2023gligen}. Specifically, we finetune GLIGEN for 180K steps for text grounded generation training and 50K steps for inpainting training with the batch size set to 8. During inference, we sample images using the DDIM~\cite{ddim} scheduler for 50 steps with the classifier-free guidance (CFG) set as 7.5. \textit{Synthetic images for training.} Taking data annotations of the training set as input, we generate synthetic images using the proposed generation pipeline and apply CAR and BOM to these images. Specifically, we construct two variants of synthetic images, named Xsyn-M and Xsyn-A, respectively. For both Xsyn-M and Xsyn-A generations, we filter out the bounding boxes smaller than a threshold ratio of the image area, and the threshold ratio is 0.1\%, 0.4\%, and 0.5\% for PIDray, OPIXray, and HiXray, respectively. For Xsyn-A generation, we generate the prohibited item in a random idle region from the background. The class of the generated item is determined by the area of the idle region to avoid the mismatch between the object size and the actual area. To deal with the above issue, we split all classes into three groups corresponding to an area interval by calculating the average area for each class, where the average area for each class is defined as the average area of all objects with the same class (see Table~\ref{tab:suppl_category_group}
for the details of class groups of each dataset). We use IoU to measure the distance between two bounding boxes, and the threshold is set to 0.2 in Eq. 3. We adopt the ViT-H SAM~\cite{kirillov2023SAM} model throughout our experiments. To prevent the disparities in generated data from affecting the model's generalization on real data, we combine the generated images with real images as the final training set, as adopted in DetDiffusion \cite{wang2024detdiffusion}. The spatial resolution of synthetic images is 512$\times$512. \textit{Detection. }We use MMDetection~\cite{mmdetection} to train downstream detectors. DINO~\cite{zhang2022dino} detector with ResNet-50 backbone is used to evaluate the dataset following the default DINO configuration of MMDetection. 
For all detectors, we uniformly train them for 6 epochs. 4 NVIDIA RTX 3090 GPUs are used for all experiments. 

\textbf{Evaluation Metrics.} 
Mean average precision (mAP), as the common metric in object detection tasks~\cite{coco}, is used to evaluate the performance. We also evaluate AP for each category and for different occlusion levels on PIDray.

\subsection{Main Results} 
In this section, we evaluate the performance of the proposed synthesis method for object detection training by supplementing real images with synthetic images generated by our method. To this end, we first compare our approach with previous methods on the PIDray dataset, and then investigate the potential of synthetic data by varying the amount of real images. Finally, we test the effectiveness of our method across various X-ray security datasets and detectors.

\begin{table*}[t!]
    \caption{Comparisons on PIDray dataset. We compare our approach with previous synthesis methods using DINO with ResNet-50 backbone on the PIDray dataset. `Easy', `Hard', and `Hidden' refer to different levels of detection difficulty. `BA', `PL', `HM', `PO', `SC', `WR', `GU', `BU', `SP', `HA', `KN' and `LI' suggest Baton, Pliers, Hammer, Powerbank, Scissors, Wrench, Gun, Bullet, Sprayer, HandCuffs, Knife and Lighter, respectively. *: represents the original L2I generation setting.}
    \label{tab:main_result}
    \centering
\tablestyle{4pt}{1.2}\scriptsize
\resizebox{1.0\textwidth}{!}{
    \begin{tabular}{l|ccccc|cccccccccccc}
    \toprule
    \multirow{2}{*}{Method} & \multicolumn{16}{c}{Average Precision$\uparrow$}                                       \\ \cline{2-18} 
                            & mAP & AP$_{50}$   & Easy   & Hard   & Hidden  & BA   & PL   & HM   & PO   & SC   & WR   & GU   & BU   & SP   & HA   & KN   & LI   \\ \midrule
    Real only               & 68.4 & 81.7  & 74.0   & 69.7   & 52.1    & 76.2 & 86.1 & 83.9 & 74.8 & 72.1 & 90.6 & 29.6 & 62.2 & 56.2 & 89.6 & 38.7 & 61.0 \\ \midrule
    \textcolor{gray}{TIP}~\cite{bhowmik2019good}  & \textcolor{gray}{69.0} & \textcolor{gray}{82.0} & \textcolor{gray}{74.9} & \textcolor{gray}{70.9} &  \textcolor{gray}{51.1}  & \textcolor{gray}{75.9} & \textcolor{gray}{86.4} & \textcolor{gray}{84.0} & \textcolor{gray}{74.7} & \textcolor{gray}{74.5} & \textcolor{gray}{91.4} & \textcolor{gray}{27.4} & \textcolor{gray}{63.2} & \textcolor{gray}{59.2} & \textcolor{gray}{89.5} & \textcolor{gray}{43.1} & \textcolor{gray}{58.8} \\
    
    \textcolor{gray}{CT-GAN}~\cite{wei2018ct-gan} & \textcolor{gray}{69.4} & \textcolor{gray}{82.4} & \textcolor{gray}{75.3} & \textcolor{gray}{71.0} & \textcolor{gray}{52.1} & \textcolor{gray}{75.9} & \textcolor{gray}{86.4} & \textcolor{gray}{83.7} & \textcolor{gray}{74.0} & \textcolor{gray}{73.2} & \textcolor{gray}{91.8} & \textcolor{gray}{35.4} & \textcolor{gray}{62.2} & \textcolor{gray}{59.3} & \textcolor{gray}{90.2} & \textcolor{gray}{39.5} & \textcolor{gray}{60.8} \\

    \textcolor{gray}{SAGAN}~\cite{zhang2019sagan} & \textcolor{gray}{69.5} & \textcolor{gray}{82.2} & \textcolor{gray}{75.0} & \textcolor{gray}{70.9} & \textcolor{gray}{53.5} & \textcolor{gray}{76.2} & \textcolor{gray}{88.1} & \textcolor{gray}{85.0} & \textcolor{gray}{75.2} & \textcolor{gray}{74.5} & \textcolor{gray}{91.7} & \textcolor{gray}{29.6} & \textcolor{gray}{62.5} & \textcolor{gray}{61.7} & \textcolor{gray}{89.8} & \textcolor{gray}{40.7} & \textcolor{gray}{59.5} \\
     \midrule
    GeoDiffusion~\cite{chen2023geodiffusion}  &   64.6 & 78.4  & 71.6  & 64.6 & 47.6 &  72.6    & 82.2     &   78.8   &   73.6   &  69.8    &  88.1    &  25.1    &  57.5    &   56.2   &   86.7   &   28.0   &   56.6   \\ 
    GLIGEN*~\cite{li2023gligen}   & 64.9 & 78.6  & 73.1  & 65.2   & 45.3   & 72.0 & 83.2 & 76.6 & 71.4 & 69.4 & 88.0 & 28.0 & 57.6 & 55.6 & 88.4 & 32.5 & 56.1 \\
    \rowcolor{red!10}
    \textbf{Xsyn-M}   & 69.1 & 82.1 & 75.5  & 70.8  &50.7  & 73.4  &  86.5  &  84.2  & 75.8 & 72.9  &   91.0   &  35.5   &   63.6   &   60.2   &  89.8  &  36.1  &   60.0   \\
    \rowcolor{red!10}
    \textbf{Xsyn-A}  & \textbf{70.7} & \textbf{83.8} & \textbf{76.8}  & \textbf{71.7}  & \textbf{54.1}  & \textbf{76.7} & \textbf{85.6}   & \textbf{85.1} &  \textbf{76.1}  & \textbf{74.8} & \textbf{91.7}  &  \textbf{36.8} &   \textbf{64.1} &  \textbf{63.5} & \textbf{89.2} &  \textbf{44.5}  &  \textbf{60.1}    \\
     \bottomrule
    \end{tabular}
}
\end{table*}

\begin{figure}
    \centering
    \includegraphics[width=1\linewidth]{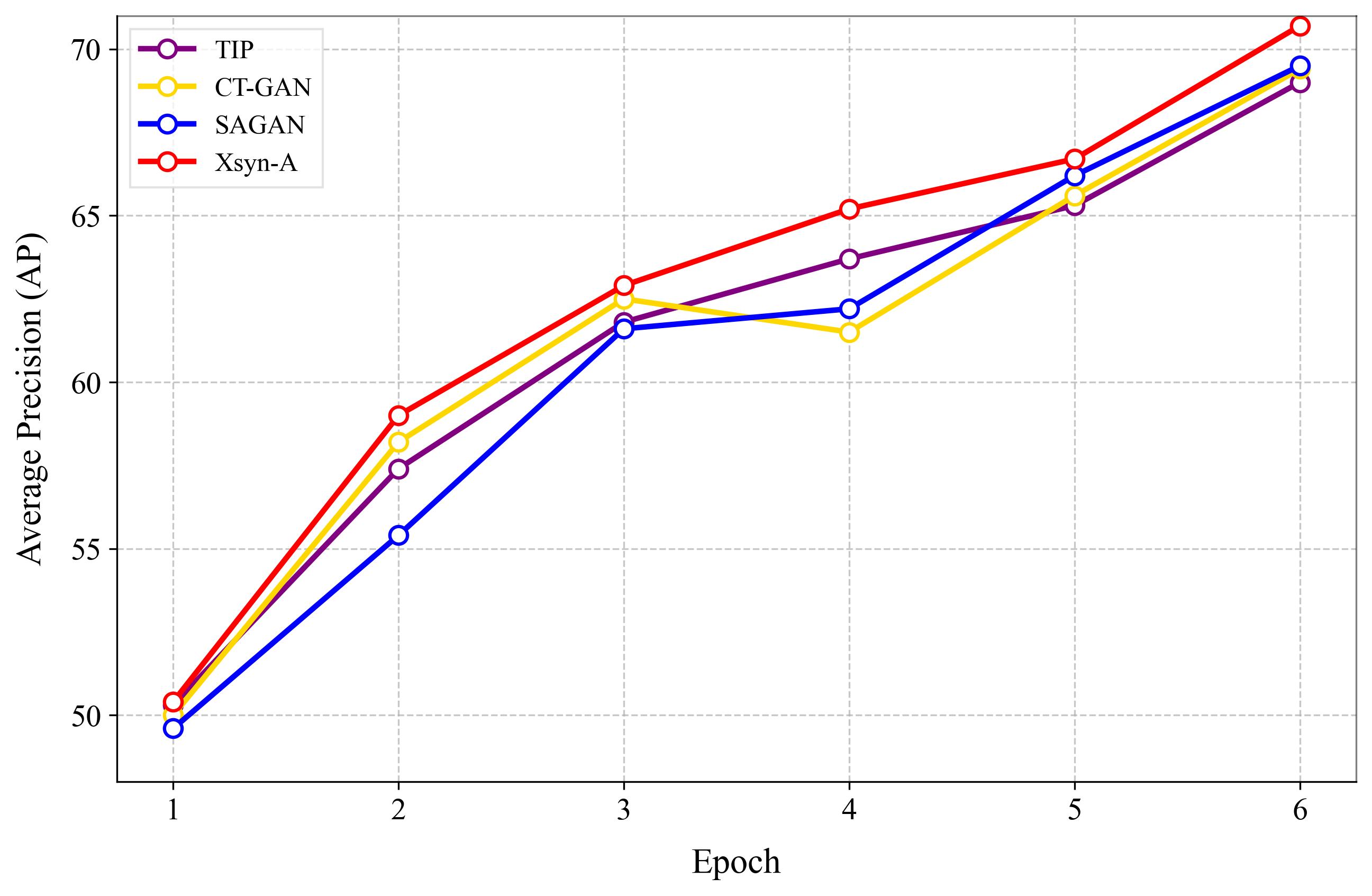}
    \caption{Potential of synthetic data. Our synthetic data achieves the best detection performance throughout the whole training period.}
    \label{fig:real_ratio}
\end{figure}

\textbf{Setup.}
For comparison experiments, we compare Xsyn-M with synthesis methods of the natural image domain and Xsyn-A with previous labor-intensive X-ray security image synthesis methods. We train all detectors for 6 epochs and reduce the learning rate by a factor of 10 in the last epoch, following their default model and training parameters in MMDetection~\cite{mmdetection}. \textit{Comparison experiments. }We choose TIP~\cite{bhowmik2019good}, CT-GAN~\cite{wei2018ct-gan}, SAGAN~\cite{zhang2019sagan}, GLIGEN~\cite{li2023gligen} and GeoDiffusion~\cite{chen2023geodiffusion} for comparison experiments following their default parameter settings. Specifically, for X-ray image synthesis methods, we extract foreground threat images using the mask annotation of PIDray dataset to implement TIP, CT-GAN, and SAGAN. We follow Yang~\cite{yang2019data} and Zhu~\cite{zhu2020data} to train CT-GAN and SAGAN on each class for 200K steps, with image size set to 128 and batch size set to 8. For L2I methods, we train GLIGEN for pure layout generation for 180K steps and train GeoDiffusion for 230K steps for fair comparison. The batch size is set to 8 for both methods.

\begin{table}[t]
    \centering
    \caption{Performance on OPIXray and HiXray. Our method is effective for various X-ray security datasets.}
\label{tab:across_datasets}
\tablestyle{4pt}{1.2}\scriptsize
\setlength{\tabcolsep}{8pt}
\begin{tabular}{l|l|lll}
\toprule
Dataset  & Setting & mAP & AP$_{50}$ & AP$_{75}$ \\ \midrule
\multirow{2}{*}{OPIXray~\cite{wei2020opixray}} & Real only  & 39.5  & 90.2  &  26.0    \\
                        & \cellcolor{red!10}\textbf{+Xsyn-A} & \cellcolor{red!10}\textbf{40.1} &  \cellcolor{red!10}\textbf{90.1} & \cellcolor{red!10}\textbf{26.1}          \\ \midrule
\multirow{2}{*}{HiXray~\cite{tao2021Hixray}} & Real only  & 49.3  & 83.4  & 53.2     \\
                        & \cellcolor{red!10}\textbf{+Xsyn-A} & \cellcolor{red!10}\textbf{50.4}  &  \cellcolor{red!10}\textbf{83.9}  & \cellcolor{red!10}\textbf{55.5}         \\ \bottomrule
\end{tabular}

\end{table}

\textbf{Comparisons with previous methods. }Table~\ref{tab:main_result} shows the results of object detection on PIDray dataset. Our Xsyn-M achieves superior performance compared with methods in the natural image domain, revealing the advantages of the proposed synthesis pipeline. Besides, Xsyn-M can achieve a competitive performance, \ie, 69.1\% $v.s.$ 69.5\% for mAP compared with SAGAN~\cite{zhang2019sagan}, and Xsyn-A can further surpass it by 1.2\% mAP. It is worth noting that our synthetic data does not require additional labor compared with previous methods, while data produced by TIP~\cite{bhowmik2019good}, CT-GAN~\cite{wei2018ct-gan}, and SAGAN~\cite{zhang2019sagan} rely on laborious pixel-wise foreground extraction. Both Xsyn-M and Xsyn-A show consistent improvement for almost all classes, especially for some difficult classes (\eg,  \textbf{+7.2\%} for Gun with Xsyn-A).

\textbf{Potential of synthetic data. }As shown in Figure~\ref{fig:real_ratio}, we plot the validation mAP curve on PIDray, and the synthetic data generated by our method has better training efficiency compared with previous methods. It indicates that our synthetic data has learned the distribution of X-ray prohibited items and can lead a faster training convergence.

\begin{table}[t]

    \centering
    \caption{Performance on various detectors. Our method can improve prohibited item detection performance consistently, regardless of detectors and backbone architectures.}
\label{tab:across_detectors}
\tablestyle{4pt}{1.2}\scriptsize
\setlength{\tabcolsep}{6pt}
\begin{tabular}{llll|lll}
\toprule
Type                          & Stage                  & Method & Backbone & mAP & AP$_{50}$ & AP$_{75}$ \\ \midrule
\multirow{6}{*}{CNN-based}    & \multirow{2}{*}{one}   & ATSS~\cite{zhang2020atss}   & R101 & 65.2 & 80.8 & 72.6       \\
                              &                        & \cellcolor{red!10}\textbf{+Xsyn-A} & \cellcolor{red!10}\textbf{R101} & \cellcolor{red!10}\textbf{65.5} & \cellcolor{red!10}\textbf{81.3} & \cellcolor{red!10}\textbf{73.0} \\ \cmidrule{2-7} 
                              & \multirow{4}{*}{two}   & C-RNN~\cite{cai2018c-rcnn}  & R101     & 68.0 & 82.6 & 75.5 \\
                              &                        & \cellcolor{red!10}\textbf{+Xsyn-A}  & \cellcolor{red!10}\textbf{R101}  & \cellcolor{red!10}\textbf{69.1} & \cellcolor{red!10}\textbf{83.4}  & \cellcolor{red!10}\textbf{76.4}  \\
                              &                        & C-RNN~\cite{cai2018c-rcnn}  & X101 & 69.6 & 83.7 & 77.0      \\
                              &                        & \cellcolor{red!10}\textbf{+Xsyn-A}  & \cellcolor{red!10}\textbf{X101} & \cellcolor{red!10}\textbf{70.2}  & \cellcolor{red!10}\textbf{84.3} & \cellcolor{red!10}\textbf{77.4}   \\ \midrule
\multicolumn{2}{l}{\multirow{4}{*}{Transformer-based}} & DINO   & R50  &  68.4  & 81.7 & 73.5  \\
\multicolumn{2}{l}{}                                   & \cellcolor{red!10}\textbf{+Xsyn-A}  & \cellcolor{red!10}\textbf{R50} & \cellcolor{red!10}\textbf{70.7} & \cellcolor{red!10}\textbf{83.8} & \cellcolor{red!10}\textbf{76.7}        \\
\multicolumn{2}{l}{}                                   & DINO   & Swin     &  76.1 & 88.6 & 81.8  \\
\multicolumn{2}{l}{}                                   & \cellcolor{red!10}\textbf{+Xsyn-A}  & \cellcolor{red!10}\textbf{Swin} & \cellcolor{red!10}\textbf{78.1}  & \cellcolor{red!10}\textbf{89.9} &  \cellcolor{red!10}\textbf{83.5}
\\ \bottomrule
\end{tabular}

\end{table}

\begin{table}[t]
    \centering
    \caption{Comparison results on OPIXray and HiXray.}
\label{tab:opi_hi_gligen}
\tablestyle{4pt}{1.2}\scriptsize
\begin{tabular}{l|ccc|ccc}
\toprule
\multirow{2}{*}{Method} & \multicolumn{3}{c|}{OPIXray} & \multicolumn{3}{c}{HiXray} \\ \cline{2-7} 
                        & mAP    & AP$_{50}$   & AP$_{75}$   & mAP   & AP$_{50}$   & AP$_{75}$  \\ \midrule
GLIGEN   & 36.7  &  88.6   &  19.1 &  49.1   &  82.0   &  53.2    \\
\rowcolor{red!10}
\textbf{Ours} & \textbf{40.1} & \textbf{90.1} & \textbf{26.1}  &    \textbf{50.4}   &  \textbf{83.9}  & \textbf{55.5}  \\ 
\bottomrule
\end{tabular}

\end{table}

\textbf{Performance on more datasets and detectors.} We extend the evaluation of our method on the OPIXray and HiXray datasets, respectively. The results in Table~\ref{tab:across_datasets} demonstrate that our method improves detection performance across various datasets. 
We further conduct experiments on various detectors, including CNN-based and Transformer-based~\cite{vaswani2017transformer} architectures, to evaluate the generalization ability. As shown in Table~\ref{tab:across_detectors}, our synthetic images achieve consistent improvement regardless of the detection models. 

\textbf{Comparison on other datasets. }
Since OPIXray and HiXray datasets lack mask annotations and FTI images, we cannot implement two-stage methods on them. Table~\ref{tab:opi_hi_gligen} shows the comparison between our method and GLIGEN, and the result demonstrates the superiority of our method. 

\subsection{Ablation Study} 

In this section, we conduct ablation studies on the proposed strategies and their specific design choices, respectively. We first ablate the parameter setting of CAR, and then ablate BOM on the basis of CAR.
All ablation studies are conducted on the best-performing Xsyn-A, and PIDray dataset is used for all experiments.

\textbf{The proposed strategies. }Table~\ref{tab:ablation_strategy} presents the impact of the proposed strategies. We analyze the effect of each proposed strategy by sequentially adding 1) CAR and 2) BOM. The results demonstrate the relative importance of each strategy, with all strategies performing the best.

\begin{table}[t]
\caption{Ablation studies on proposed strategies. We first add CAR and then BOM to investigate their performance separately. Best results are achieved when both strategies are adopted.}
    \label{tab:ablation_strategy}
    \centering
    \small
    \tablestyle{4pt}{1.2}\scriptsize
\setlength{\tabcolsep}{8pt}
\begin{tabular}{l|lll}
\toprule
   Method & \multicolumn{1}{c}{mAP} & \multicolumn{1}{c}{AP$_{50}$} & \multicolumn{1}{c}{AP$_{75}$} \\ \midrule
Real only    &  68.4   &   81.7  &   73.9  \\
+Xsyn-A (w/o CAR)  &   69.6  &  82.3   &   75.5  \\
+Xsyn-A (w/o BOM)  &  70.3   &   83.1  &  76.0   \\
\rowcolor{red!10}
\textbf{+Xsyn-A} &    \textbf{70.7}  &   \textbf{83.8}  &   \textbf{76.7}    \\

 \bottomrule
\end{tabular}
\end{table}

\textbf{Hyper-parameters of proposed strategies. }
\textit{Median point sampling. }Table~\ref{tab:ablation_param} (upper) shows the performance of CAR using different division times $n$, where $n=0$ means that we only use the grounding box to implement refinement. The gain reaches its biggest when $n=4$, indicating the benefit of incorporating median points and suggesting that MPS has good scalability for annotation refinement. We set $n$ to 4 for other experiments. 
\textit{Latent occlusion coefficient. }Table~\ref{tab:ablation_param} (bottom) provides the ablation study for occlusion coefficient $\alpha$. The performance increases when $\alpha$ changes from 0.1 to 0.3, while it decreases from 0.3 to 0.7. The result suggests that a medium occlusion coefficient is beneficial to enhance the imaging complexity, while a too small or too large occlusion coefficient cannot model the complex occlusion in real-world baggage. Therefore, the optimum $\alpha$ is set to 0.3 for better imaging complexity enhancement.

\begin{table}[t]\label{tab:ablation_param}
    \centering
    \caption{Ablations on hyper-parameters of proposed strategies. We ablate the choice of division times \textit{n} for CAR and latent occlusion coefficient $\alpha$ for BOM respectively on Xsyn-A.}
\label{tab:ablation_param}
\tablestyle{4pt}{1.2}\scriptsize
\setlength{\tabcolsep}{9pt}
\begin{tabular}{l|l|lll}
\toprule
Type  & Setting & mAP & AP$_{50}$ & AP$_{75}$ \\ \midrule

\multirow{5}{*}{CAR-$n$} & 0  &  69.7   &   82.5   &  75.6  \\
                        & 1  & 69.9  & 82.7 & 75.9  \\
                        & 2  &  70.1  & 83.0  & 75.9      \\
                        & 3  &  70.2  & 82.8  &   76.0      \\
                        & \cellcolor{red!10}\textbf{4}  &  \cellcolor{red!10}\textbf{70.3}  &  \cellcolor{red!10}\textbf{83.1}  & \cellcolor{red!10}\textbf{76.0}         \\ \midrule
\multirow{4}{*}{BOM-$\alpha$} & 0.1  &  70.3   &  83.1  &  76.3    \\
& \cellcolor{red!10}\textbf{0.3}  &  \cellcolor{red!10}\textbf{70.7}  & \cellcolor{red!10}\textbf{83.8}   &  \cellcolor{red!10}\textbf{76.7}     \\
                        & 0.5  &  70.2  &  82.8  &    76.0     \\
                        
                        & 0.7  &  69.8   &  82.5  &   75.5       \\ \bottomrule
\end{tabular}
\end{table}

\begin{table}[t]
\caption{Study on point sampling strategies in CAR.}
    \label{tab:sampling_strategy}
    \centering
    \tablestyle{4pt}{1.2}\scriptsize
\setlength{\tabcolsep}{10pt}
\begin{tabular}{l|lll}
\toprule
Point Sampling & \multicolumn{1}{c}{mAP} & \multicolumn{1}{c}{AP$_{50}$} & \multicolumn{1}{c}{AP$_{75}$} \\ \midrule
Top-1    & 69.8     &   82.7   &  75.8   \\
Top-15    &  69.8  &  82.9  &  75.7   \\ \midrule
MPS (\textit{n} = 1) & 69.9  & 82.7 & 75.9 \\
\rowcolor{red!10}
\textbf{MPS (\textit{n} = 4)} &    \textbf{70.3}  &   \textbf{83.1}  & \textbf{76.0}    \\

 \bottomrule
\end{tabular}
\end{table}

\begin{table}[t]
\centering
\small
\caption{BOM ablations on occlusion space and period.}

\label{tab:ablation_blom_latent_pixel}
\tablestyle{4pt}{1.2}\scriptsize
\setlength{\tabcolsep}{7pt}
\begin{tabular}{l|c|ccc}
\toprule
                              & Period & mAP & AP$_{50}$ & AP$_{75}$ \\ \midrule
\multirow{2}{*}{Latent Space} & $t$   & 69.9  &  82.9   &  75.5        \\
& \cellcolor{red!10}$T$  &  \cellcolor{red!10}\textbf{70.7}  &  \cellcolor{red!10}\textbf{83.8}    & \cellcolor{red!10}\textbf{76.7}     \\
                               \midrule
Pixel Space                   & -      &   69.9  &   82.7    &   75.8     \\ \bottomrule
\end{tabular}

\end{table}

\textbf{Point sampling strategies.}
We study the choice of point sampling strategies in CAR by designing a top-\textit{k} point sampling strategy for comparison. Specifically, the top-\textit{k} strategy samples \textit{k}+1 points in total, which consists of \textit{k} foreground points with top-\textit{k} activation values and one background point with the minimum activation value within the cross-attention map. We set \textit{k} to 1 and 15 to compare with MPS (\textit{n} = 1) and MPS (\textit{n} = 4), respectively.
The performance comparison between our median point sampling strategy and the top-\textit{k} point sampling strategy is presented in Table~\ref{tab:sampling_strategy}. 
The result demonstrates the superiority of our strategy, indicating that the MPS strategy has better scalability for annotation refinement. 

\textbf{Occlusion space and period. }The ablation study for occlusion space and period is shown in Table~\ref{tab:ablation_blom_latent_pixel}. We fuse the occluder region with foreground regions in the original image to implement occlusion modeling in pixel space. The result shows that modeling occlusion in latent space achieves better performance than in pixel space. We also investigate the influence of the occlusion period by modeling occlusion at each denoising step $t$, but the performance is much lower than the original version. We argue that such an approach may destroy the distribution of foregrounds in the cross-attention map, thus affecting the process of CAR.

\begin{figure*}[t]
    \centering
    \includegraphics[width=1\linewidth]{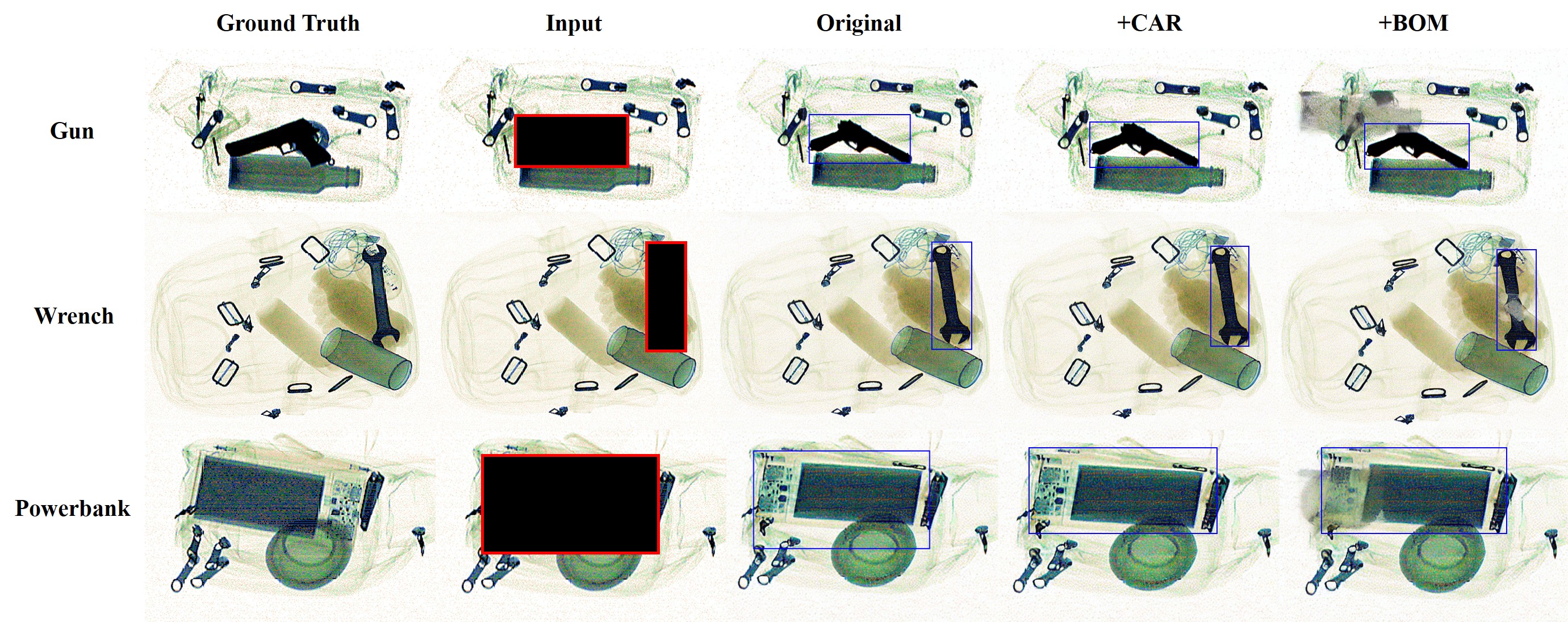}
    \caption{Qualitative results on PIDray dataset. Our method can synthesize well-annotated and realistic X-ray security images. The blue boxes in the 3rd column and the last two columns refer to the input grounding boxes and the refined annotation boxes, respectively. Please zoom in for better visualization. }
    \label{fig:qualitative_result}
\end{figure*}

\subsection{Qualitative Results}
We provide qualitative results on the PIDray dataset shown in Figure~\ref{fig:qualitative_result}. The original results in the 3rd column have obvious spatial misalignment between the generated prohibited item and the bounding box. When we apply CAR to the original results, the bounding box is refined to enclose the prohibited item tightly, as shown in the 4th column. We further enhance the imaging complexity in the 5th column by using BOM to occlude the prohibited item. It is worth mentioning that we perform CAR on the original image and apply BOM to obtain the hidden image, which ensures that the annotation refinement will not be compromised by the introduction of occlusion.

\begin{figure*}
    \centering
    \includegraphics[width=1\linewidth]{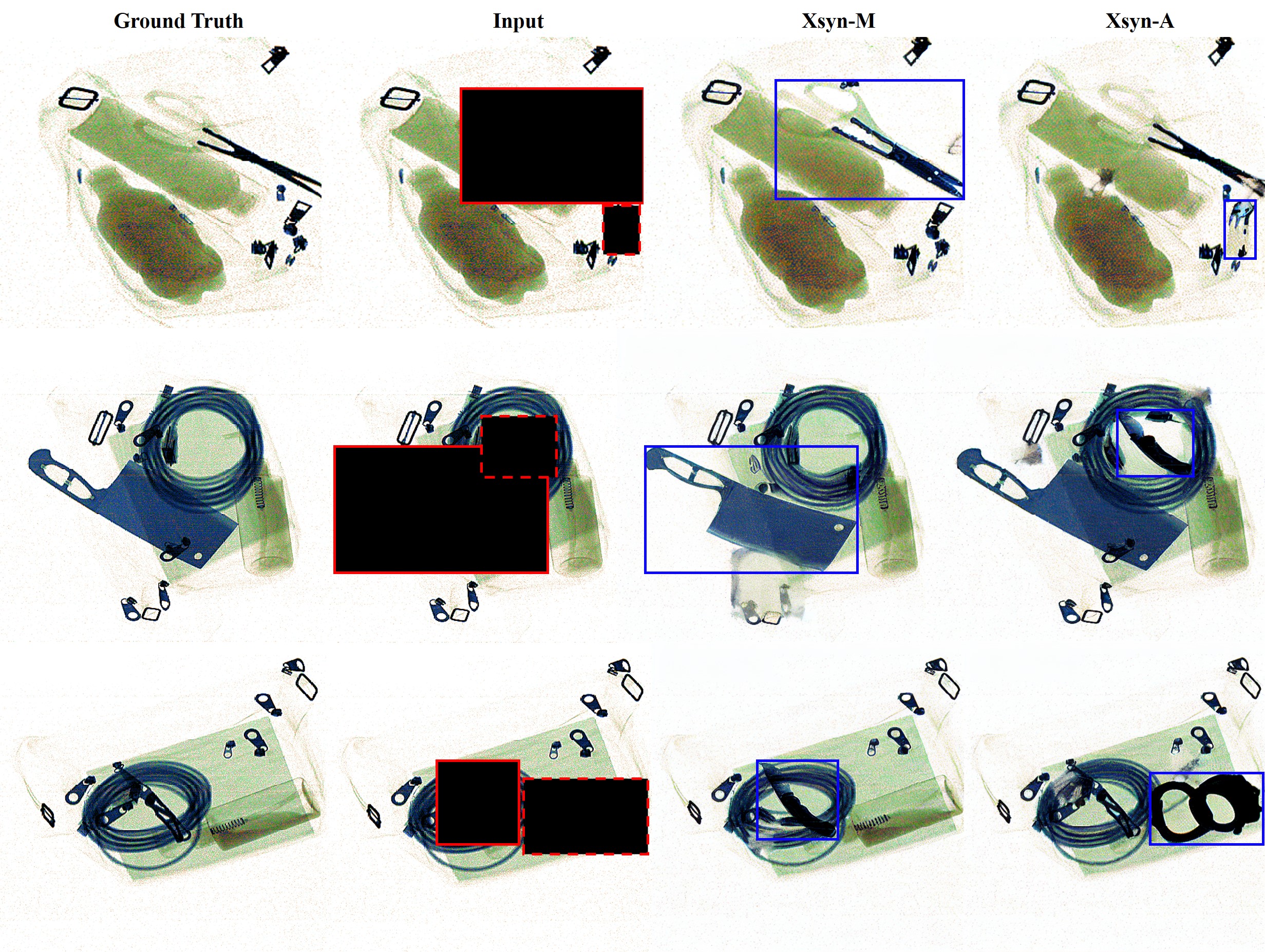}
    \caption{More qualitative results on PIDray dataset. The solid-line and dashed-line boxes in the 2nd column are grounding boxes for Xsyn-M and Xsyn-A generation, respectively. From top to bottom, the text prompt for Xsyn-M in each row is "Scissors", "Knife", and "Knife", and the text prompt for Xsyn-A in each row is "Lighter", "Knife", and "HandCuffs", respectively.}
    \label{fig:vis_syn_1_2}
\end{figure*}

Figure~\ref{fig:vis_syn_1_2} presents more visualization of X-ray security images synthesized by our approach. The synthetic images from Xsyn-A can effectively simulate the scenario of concealing contraband in real-world baggage, which is especially well illustrated by the second image in the last column. 

\begin{figure*}
    \centering
    \includegraphics[width=1\linewidth]{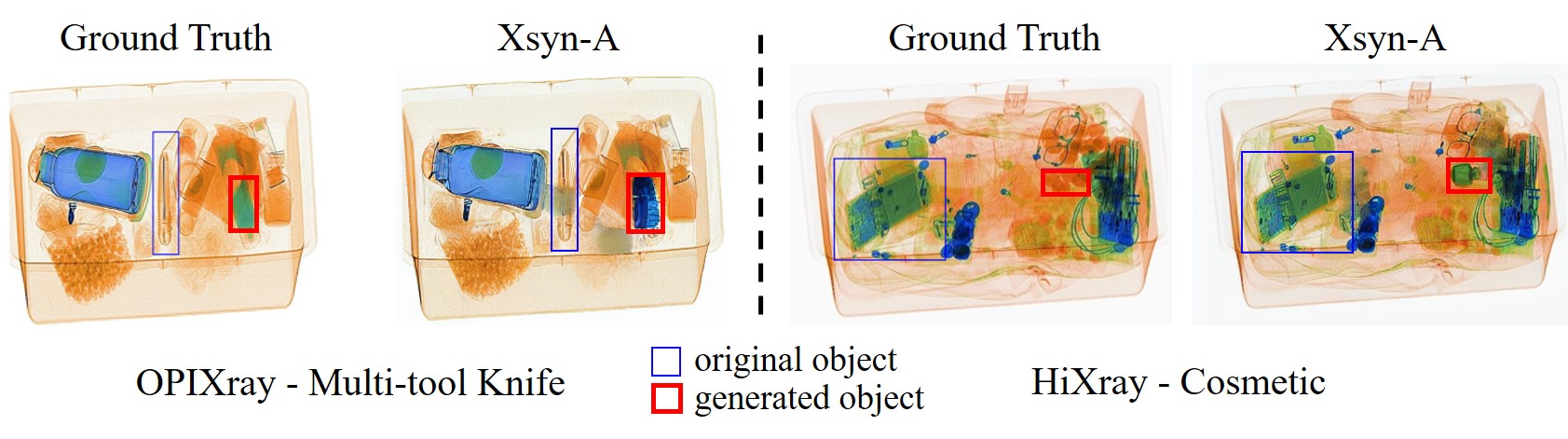}
    \caption{More qualitative results on OPIXray and HiXray dataset. The blue box refers to the original object, and the red box refers to the generated object using our method.}
    \label{fig:opi_hi_qual}
\end{figure*}

Figure~\ref{fig:opi_hi_qual} further shows qualitative results for OPIXray and HiXray datasets, demonstrating that our method can synthesize realistic objects across different datasets.

\section{Discussions}
\label{sec:more_discuss}
Our method can be adapted to other domains (\eg, remote sensing) where data collection and annotation are hard. With the great development of security inspection equipment, Computed Tomography (CT) has emerged as a more advanced imaging modality for contraband detection. However, compared with 2D X-ray security images, collecting and annotating CT images is more challenging due to their 3D volumetric characteristics. Therefore, it is a promising direction to explore CT image synthesis using generative models and exploit the synergy between X-ray and CT imaging modalities.
We hope that this paper encourages further research on security image (\eg, X-ray and CT) generation to benefit automatic prohibited item detection. 

\section{Conclusion}
In this paper, we propose Xsyn, a simple and effective one-stage X-ray security image synthesis pipeline to generate high-quality prohibited item detection data. In contrast to the previous two-stage methods, for the first time, our method removes the labor-intensive foreground extraction procedure. To improve the usability of generative synthetic data, 
our method incorporates two effective strategies to automatically refine the synthetic annotation and enhance the synthetic complexity. The synthetic images generated by our method can improve the prohibited item detection performance across various public datasets and detectors. We hope Xsyn can bring new inspiration for exploiting the potential of generative synthetic data in the X-ray security domain.


\bibliographystyle{IEEEtran}
\bibliography{ieee}









\end{document}